\documentclass[10pt,twocolumn,letterpaper]{article}

\usepackage{3dv}
\usepackage{times}
\usepackage{epsfig}
\usepackage{graphicx}
\usepackage{amsmath}
\usepackage{amssymb}
\usepackage{stfloats}

\usepackage{wrapfig}

\usepackage[pagebackref=true,breaklinks=true,colorlinks,bookmarks=false]{hyperref}
\usepackage{multirow} 
\usepackage{booktabs}

\threedvfinalcopy 

\def\threedvPaperID{****}

\ifthreedvfinal\fi
\begin{document}

\title{RAFT-Stereo: Multilevel Recurrent Field Transforms for Stereo Matching}

\author{Lahav Lipson\\
Princeton University\\

\and
Zachary Teed\\
Princeton University\\

\and
Jia Deng\\
Princeton University\\
}

\maketitle

\begin{abstract}
   We introduce RAFT-Stereo, a new deep architecture for rectified stereo based on the optical flow network RAFT~\cite{teed2020raft}. We introduce multi-level convolutional GRUs, which more efficiently propagate information across the image. A modified version of RAFT-Stereo can perform accurate real-time inference. RAFT-stereo ranks first on the Middlebury leaderboard, outperforming the next best method on 1px error by $29\%$ and outperforms all published work on the ETH3D two-view stereo benchmark. Code is available at \url{https://github.com/princeton-vl/RAFT-Stereo}.
\end{abstract}

\section{Introduction}
Stereo depth estimation is a fundamental vision problem with direct applications in robotics, augmented reality, photogrammetry, and video understanding problems. In the standard setup, two frames---a left frame and a right frame---are provided as input. The task is to estimate a pixelwise displacement map between the input images. In rectified stereo, the displacement of each pixel is constrained to a horizontal line. This displacement map, termed disparity, can be used alongside camera calibration parameters to recover depth, a 3D point cloud, or other 3D representations suitable for the target downstream application.

Early work has focused on two key parts of the problem: (1) feature matching and (2) regularization. Given two images, feature matching aims to compute a matching cost between a pair of image patches. Commonly used methods include mutual information \cite{hirschmuller2007stereo}, normalized cross-correlation \cite{heo2010robust}, and the census transform followed by Hamming distance \cite{fife2012improved}.  Given a set of noisy matches, regularization aims to recover a consistent depth map subject to priors such as smoothness and planarity. These two objectives can be naturally formulated as an optimization problem, maximizing some measure of visual similarity subject to priors over 3D geometry.

Optical flow and rectified stereo are closely related problems. In optical flow, the task is to predict a pixelwise displacement field, such that for every pixel in the first frame, we can estimate its correspondence in the second frame. In rectified stereo, the task is the same, except that we have the additional constraints that the x-displacement is always positive and the corresponding points lie on a horizontal line---hence, the y-displacement is always 0. 

Despite the similarities between stereo and flow, neural network architectures for the two tasks are vastly different. In stereo, the predominant approach has been the use of 3D convolutional neural networks. First a 3D cost volume is built by enumerating integer disparities, then use a 3D convolutional network to filter the cost volume \cite{mvnsnet,psmnet,gcnet,gwcnet,ganet,dsmnet}. This formulation leverages stereo geometry as an inductive prior in network design. However, using 3D convolutions to process the cost volume comes at a high computational cost and limits the possible operating resolution. Specialized approaches are required to operate at high resolutions~\cite{yang2019hierarchical} such as the mega-pixel images from the Middlebury dataset~\cite{middlebury}.

On the other hand, optical flow is typically approached using iterative refinement. RAFT~\cite{teed2020raft} showed that iterative refinement can be performed entirely at high resolution, proposing a simple architecture that performed well on standard flow benchmarks. RAFT first extracts features from the input images, then builds a 4D cost volume by computing the correlation between all pairs of pixels. Finally, a GRU-based update operator iteratively updates the flow field using features retrieved from the correlation volume. 

\begin{figure*}[h]
    \centering
	\includegraphics[width=.99\textwidth]{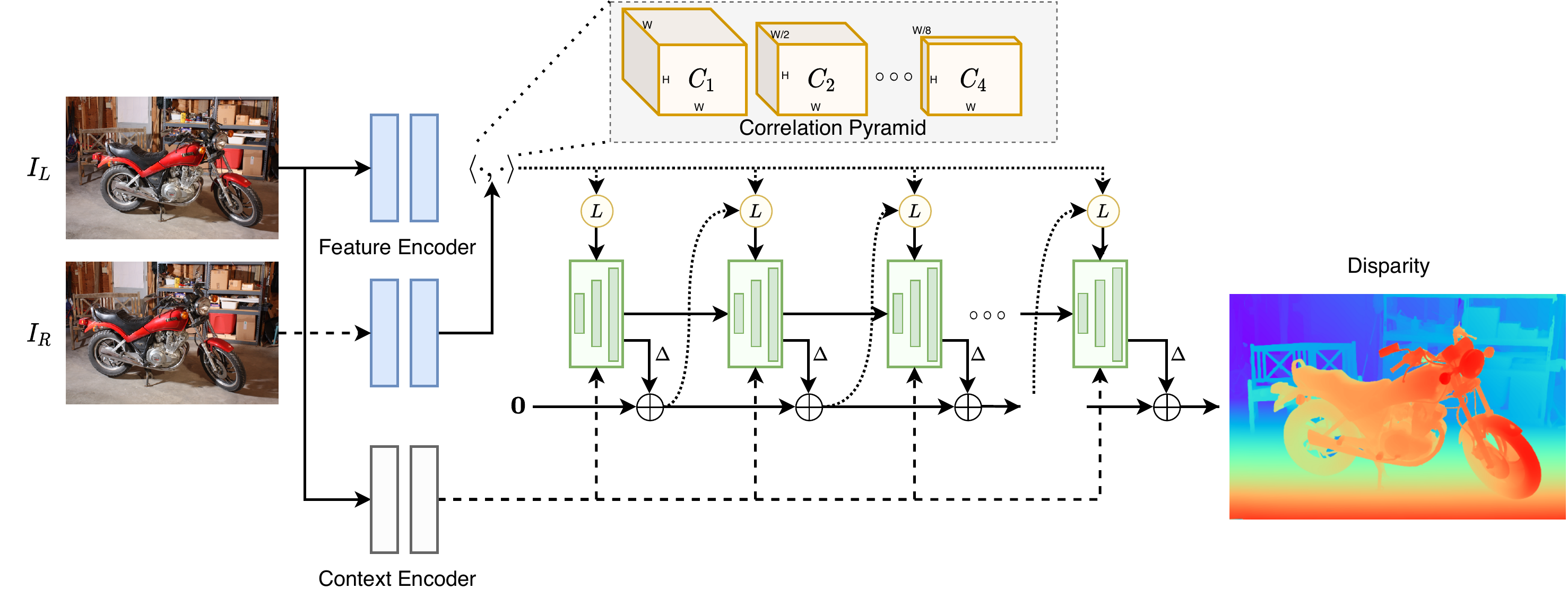}
	\caption{Correlation features (blue) are extracted from each of the images and are used to construct the correlation pyramid. "Context" image features (white) and an initial hidden state are also extracted from the context encoder. The disparity field is initialized to zero. Every iteration, the GRU(s) (green) use the current disparity estimate to sample from the correlation pyramid. The resulting correlation features, initial image features and current hidden state(s) are used by the GRU(s) to produce a new hidden state and an update to the disparity. }
	\label{fig:archdiagram}
\end{figure*}

We introduce RAFT-Stereo, a new architecture for two-view stereo. An overview of our approach is shown in Fig.~\ref{fig:archdiagram}. The overall design is based on RAFT~\cite{teed2020raft}. First, we replace the all-pairs 4D correlation volume with a 3D volume by only computing the visual similarly between pixels of the same height. Additionally, we introduce multi-level GRU units that maintain hidden states at multiple resolutions with cross-connections but still generate a single high-resolution disparity update. This improves the ability of the update operator to propagate information across the image, improving the global consistency of the disparity field.

RAFT-Stereo is substantially different from previous stereo networks. Existing work has commonly relied on 3D convolution networks to process stereo cost volumes~\cite{mvnsnet,psmnet,gcnet,gwcnet,ganet,dsmnet}. In contrast, RAFT-Stereo uses only 2D convolutions and a lightweight cost volume constructed using a single matrix multiplication. By avoiding the high computation and memory cost of 3D convolutions, RAFT-Stereo can be directly applied megapixel images without the need for resizing or processing the image in patches.  Furthermore, by using an iterative network, we can easily trade accuracy for efficiency with early stopping. RAFT-Stereo also doesn't require additional complex loss terms, making it easy to train.

Our main contribution is a new stereo network which unifies stereo and optical flow approaches. RAFT-Stereo shows much better cross-dataset generalization than existing neural networks. When trained only on synthetic data, our network performs very well on real datasets such as KITTI~\cite{kitti}, ETH3D~\cite{eth3d}, and Middlebury~\cite{middlebury}, outperforming all other works evaluated in the same setting. Additionally, RAFT-stereo is accurate. It ranks first on the Middlebury leaderboard~\cite{middlebury} and outperforms all published work on the ETH3D leaderboard~\cite{eth3d}. Due to its high accuracy and good generalization, we believe RAFT-Stereo will be useful as an off-the-self stereo algorithm.

\section{Related Work}

The task of predicting disparity between rectified stereo images is a longstanding problem in computer vision. Early work focused on designing better matching costs \cite{hannah1974computer,zabih1994non} and efficient inference algorithms \cite{kolmogorov2006convergent,semiglobalmatching,barnes2009patchmatch}. Traditional stereo pipelines generally consisted of a matching stage and a filtering stage. In the matching stage, pairwise costs were computed between images patches. In the optimization and filtering stages, priors could be imposed to correct erroneous matching and recover a consistent disparity map.

Deep learning was first applied to improve matching costs in the stereo pipeline. \v{Z}bontar and LeCun \cite{zbontar2015computing} proposed a network for evaluating a matching score between a pair of image matches. The matching costs were then processed using semiglobal matching, consistency checking, and filtering. Mayer et al. \cite{sceneflow} proposed the first end-to-end trainable stereo matching network, based on the Flownet architecture \cite{dosovitskiy2015flownet}, in addition to a large synthetic dataset which made training convolutional networks for stereo possible.

Inspired by the classical pipeline, many works have adopted a 3D neural network architecture for end-to-end stereo matching\cite{mvnsnet,psmnet,gcnet,gwcnet,ganet,dsmnet}. GCNet\cite{gcnet} was one of the first papers to propose this approach. In this framework, images are first mapped through a 2D convolutional network to obtain a dense feature representation. Next, a 3D cost volume is constructed over the 2D feature maps, either through concatenation\cite{gcnet} or correlation operator\cite{gwcnet}. The cost volume is then filtered through a series of 3D convolutional layers, before being mapped to a pointwise depth estimate through a differentiable arg-min operator. Many variations on this design have been proposed, such as using a stacked 3D hourglass to process the cost volume\cite{psmnet}, or designing new aggregation layers to better propagate information\cite{ganet}. The 3D convolutions aim to act as a differentiable approximation to classical filtering algorithms such as SGM\cite{semiglobalmatching}. 

While this approach has outperformed traditional methods such on datasets such as KITTI\cite{kitti} and FlyingThings3D\cite{sceneflow} the 3D convolutions come at a high computational cost and often fail to generalize outside the domain they were trained, meaning that they cannot be readily used on datasets which don't have ground truth training data. There have been several efforts to improve the generalization ability of deep stereo networks such as the addition of new network components\cite{dsmnet} or generating additional training data \cite{watson2020learning}. DSMNet \cite{dsmnet} tries to improve the generalization ability of the GA-Net architecture by normalizing the features used to construct the cost volume and by utilizing a non-local graph-based filtering approach which reduces GA-Net's dependence on local patterns. DSMNet achieves better generalization than prior works, but still uses 3D convolutions in their architecture design. This results in a high computational cost and limits the operating resolution of DSMNet. These works have focused on zero-shot cross dataset generalization. In this paper, we also evaluate cross dataset generalization on the ETH3D\cite{eth3d}, KITTI \cite{kitti}, and Middlebury \cite{middlebury} datasets.

Another line of work has looked replacing the more costly components of the 3D networks with more lightweight modules. Liang et al. \cite{liang2018learning} first proposed a 2 stage refinement network for stereo. Bi3D \cite{badki2020bi3d} proposed estimating depth with a series of classification stages. Recently HITNet \cite{tankovich2020hitnet} leveraged the planar geometry of the scene as an inductive prior in the network design by guiding the stereo predictions using predicted tiles. In the forward pass, HITNet's tile-based method must decide if each pixel lies on a plane. To learn this behavior, they must impose several additional loss terms on the angle of the tiles and the decision weights, as opposed to RAFT-Stereo which solely uses a standard L1 loss. HITNet also maintains a running stereo prediction at full resolution, while RAFT-Stereo only upsamples the stereo prediction at the very end. This makes RAFT-Stereo more memory efficient, enabling us to predict full-resolution stereo on megapixel images.

\section{Approach}

Given a pair of rectified images $(I_L, I_R)$, we aim to estimate a disparity field $\mathbf{d}$ giving the horizontal displacement for every pixel in $I_L$. Similar to RAFT~\cite{teed2020raft} our approach is composed of three main components: a feature extractor, a correlation pyramid, and a GRU-based update operator as shown in Fig. \ref{fig:archdiagram}. The update operator iteratively retrieves features from the correlation pyramid and performs updates on the disparity field.

\subsection{Feature Extraction}
We use two separate feature extractors termed the \emph{feature encoder} and the \emph{context encoder}. The feature encoder is applied to both the left and right images and maps each image to a dense feature map, which is then used to construct the correlation volume. The network consists of a series of residual blocks and downsampling layers, producing feature maps at 1/4 or 1/8 the input image resolution with 256 channels, depending on the number of downsampling layers used in our experiments. We use instance normalization~\cite{ulyanov2016instance} in the feature encoder. 

The context encoder has identical architecture to the feature encoder except we replace instance normalization with batch normalization~\cite{ioffe2015batch} and only apply the context encoder on the left image. The context features are used to initialize the hidden state of the update operator and also injected into the GRU during each iteration of the update operator.

\subsection{Correlation Pyramid}

\noindent \textbf{Correlation Volume:} We use the dot product between feature vectors as a measure of visual similarity. Similar to how RAFT~\cite{teed2020raft} constructs a 4D correlation volume by computing the visual similarity between all pairs of pixels, we restrict computation of the correlation volume to pixels which share the same y-coordinate. Given feature maps $\mathbf{f}, \mathbf{g} \in \mathbb{R}^{H \times W \times D}$ extracted from $I_L$ and $I_R$ respectively, the 3D correlation volume can be computed using a modification of the 4D volume construction by restricting computation of the inner product to feature vectors which share the same first index:
\begin{equation}
\begin{split}
\mathbf{C}_{ijk} \ = \sum_h \mathbf{f}_{ijh} \cdot \mathbf{g}_{ikh}, \ \ \mathbf{C} \in \mathbb{R}^{H \times W \times W}
\end{split}
\end{equation}
Like the 4D volume, computation of the 3D volume can be efficiently implemented using a single matrix multiplication, which can be easily computed on the GPU and takes up only a small fraction of total runtime. 

In rectified stereo, we can typically assume that all disparities are positive; thus, the correlation volume really only needs to be computed for positive disparities. However, the advantage of computing the full volume is that the operation can be implemented using matrix multiplication which is highly optimized. This simplifies the overall architecture, allowing us to use common operations instead of requiring custom GPU kernels.

\vspace{1mm}
\noindent\textbf{Correlation Pyramid:} We construct a 4 level pyramid of correlation volumes through repeated average pooling of the last dimension. The k$^{\text{th}}$ level of the pyramid is constructed from the volume at level k using 1D average pooling with a kernel size of 2 and a stride of 2 producing a new volume $\mathbf{C}^{k+1}$ with dimension $H \times W \times W/2^k$. Each level of the pyramid has an increased receptive field, but by only pooling the last dimension, we maintain the high resolution information present in the original image, which allows us to recover very fine structures.

\vspace{1mm}

\begin{figure}[t]
    \centering
	\includegraphics[width=0.99\columnwidth]{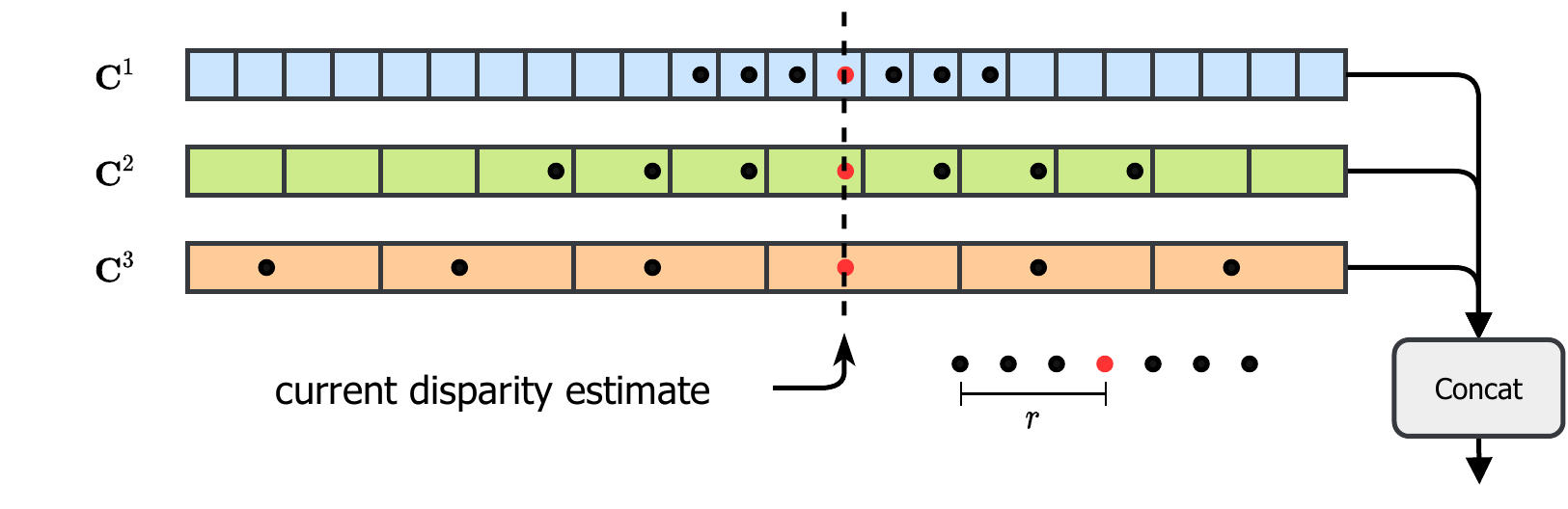}
	\caption{Lookup from the correlation pyramid. We use the current estimate of disparity to retrieve values from the each level of the correlation pyramid. We index from each level in the pyramid by linear interpolating at the current disparity estimate and at integer offsets, whose size depends on the correlation pyramid level.}\vspace{-2mm}
	\label{fig:lookup}
\end{figure}

\noindent\textbf{Correlation Lookup:} To index into the correlation pyramid, we define a lookup operator $L_{\mathbb{C}}$ analogous to the one defined in RAFT. Given a current estimate of disparity $d$, we construct a 1D grid with integer offsets around the current disparity estimate as shown in Fig.~\ref{fig:lookup}. The grid is used to index from each level in the correlation pyramid. Since grid values are real numbers, we use linear interpolation when indexing each volume. The retrieved values are then concatenated into a single feature map.

\subsection{Multi-Level Update Operator}
\label{sec:updates}

We predict a series of disparity fields $\{\mathbf{d}_1, ..., \mathbf{d}_N \}$ from an initial starting point $\mathbf{d}_0 = \mathbf{0}$. During each iteration, we use the current estimate of disparity to index the correlation volume, producing a set of correlation features. These features are passed through 2 convolutional layers. Similarly, the current disparity estimate is also passed through 2 convolutional layers. The correlation, disparity, and context features and then concatenated and injected into the GRU. The GRU updates the hidden state. The new hidden state is then used to predict the disparity update.  \smallskip

\noindent\textbf{Multiple Hidden States:} The original RAFT performs updates entirely at a fixed, high resolution. An issue with this approach is that the receptive field increases very slowly with the number of GRU updates. This can be problematic for scenes with large textureless regions with little local information. We combat this issue by proposing a multi-resolution update operator which operates on feature maps at 1/8, 1/16, and 1/32 resolutions simultaneously. In our experiments, we show that our use of a multi-resolution update operator results in better generalization performance. 

\begin{figure}[t]
    \centering
	\includegraphics[width=.99\columnwidth]{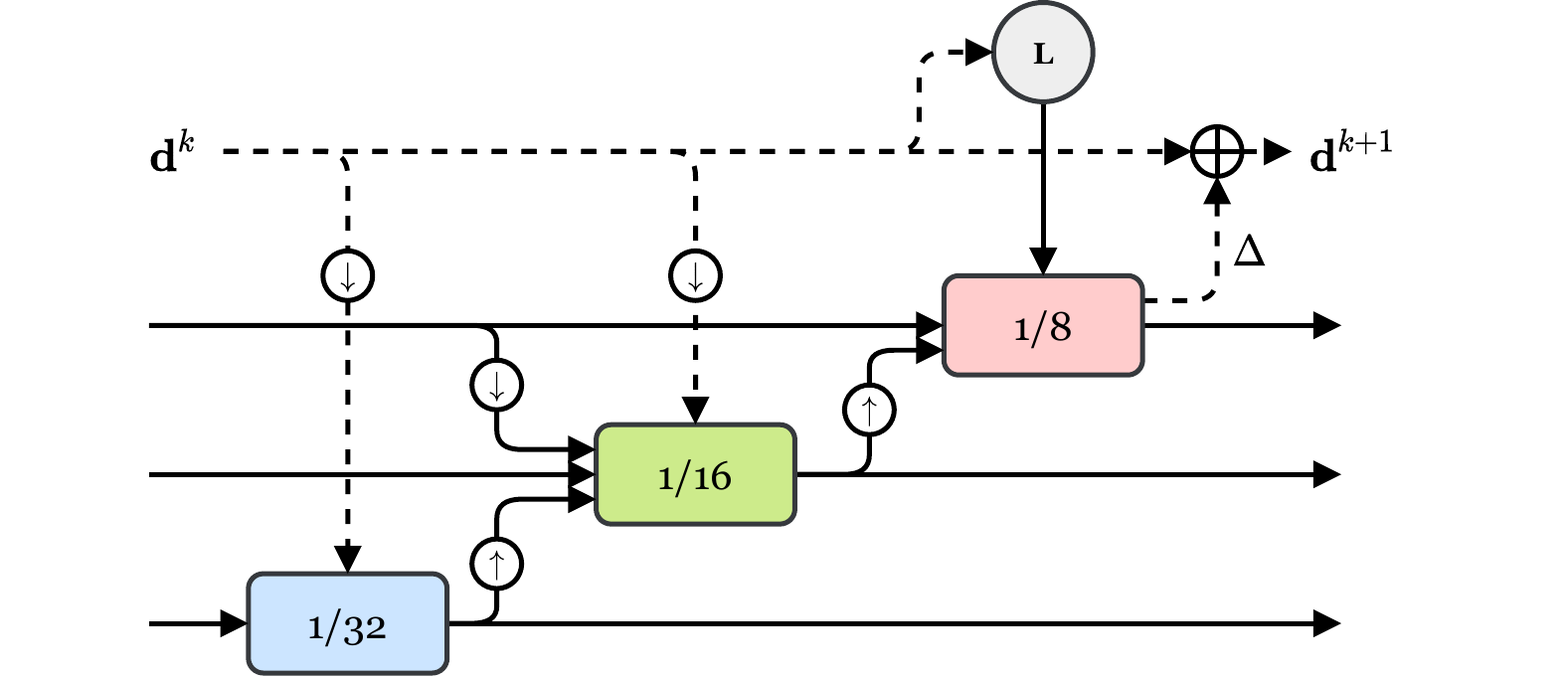}
	\caption{Multilevel GRU. We use a 3-level convolutional GRU which acts on feature maps at 1/32, 1/16, and 1/8 the input image resolution. Information is passed between GRUs at adjacent resolutions using upsampling and downsampling operations. The GRU at the highest resolution (red) performs lookups from the correlation pyramid and updates the disparity estimate.}\vspace{-0.5cm}
	\label{fig:multileveldiagram}
\end{figure}

The GRUs are \textit{cross-connected} by using each other's hidden states as input as shown in Fig.~\ref{fig:multileveldiagram}. Correlation lookup and the final disparity update is performed by the GRU at the highest resolution. We also experiment with a higher resolution model, with GRU updates at 1/4, 1/8, and 1/16 the resolution of the input image.

\noindent\textbf{Upsampling:} The predicted disparity field is at 1/4 or 1/8 the input image resolution. To output full resolution disparity maps, we use the same convex upsampling method as RAFT. RAFT-Stereo takes the full resolution disparity values to be the convex combination of the 3x3 grid of their coarse resolution neighbors. The convex combination weights are predicted by the highest resolution GRU. \\

\vspace{-4mm}
\subsection{Slow-Fast GRU}
A GRU-update to a 1/8 resolution hidden state takes approximately 4x as many FLOPs compared to updating a 1/16 resolution hidden state. In order to leverage this fact for faster inference, we train a version of RAFT-Stereo in which we update the 1/16 and 1/32 resolution hidden states several times for every single update to the 1/8 resolution hidden state. On KITTI resolution images with 32 GRU updates, this simple change reduces the runtime of RAFT-Stereo from 0.132s to 0.05s, a 52\% decrease. See table \ref{table:Ablations}.

This modification allow us to achieve performance competitive with state-of-the-art approaches for stereo vision in real-time with RAFT-Stereo (See section \ref{sec:realtimeinf}), with a method that runs an order of magnitude faster.

\subsection{Supervision} 
We supervised on the $l_1$ distance between the predicted and ground truth disparity over the full sequence of predictions, $\{\mathbf{d}_1, ..., \mathbf{d}_N \}$, with exponentially increasing weights. Given ground truth disparity $\mathbf{d}_{gt}$, the loss is defined as
\begin{equation}
    \mathcal{L} = \sum_{i=1}^{N} \gamma^{N-i} ||\mathbf{d}_{gt} - \mathbf{d}_i||_1, \qquad \text{where } \gamma=0.9 .
\end{equation}

\section{Experiments}

\begin{figure*}[h]
\centering
	\includegraphics[width=.99\textwidth]{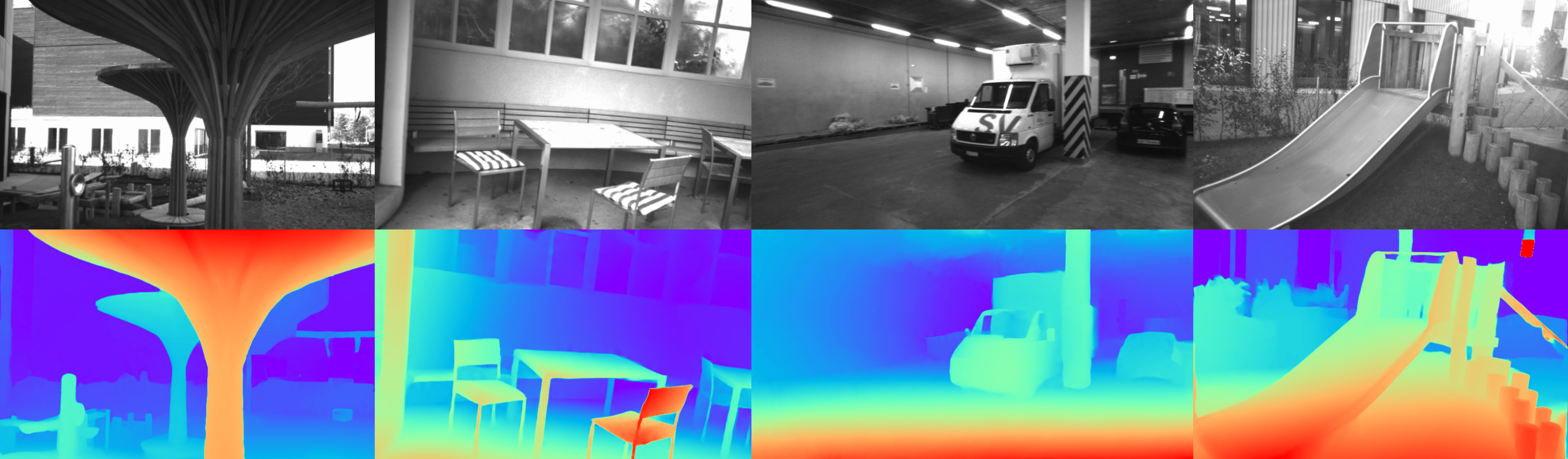}
	\caption{Results on the ETH3D stereo dataset. RAFT-Stereo is robust to difficulties like textureless surfaces and overexposure.}
	\label{fig:ETHQual}
\end{figure*}


We evaluate RAFT-Stereo on ETH3D~\cite{eth3d}, Middlebury~\cite{middlebury} and KITTI-2015~\cite{kitti}. Following previous works, we pretrain our model on the synthetic Sceneflow datasets~\cite{sceneflow}. Our method achieves state-of-the-art performance on the ETH3D and Middlebury leaderboards and we outperform existing methods in the zero-shot generalization setting on ETH3D, KITTI and Middlebury. 

\vspace{1mm}
\noindent\textbf{Implementation Details:} RAFT-Stereo is implemented in Pytorch \cite{pytorch} and is trained using two RTX 6000 GPUs. All modules are initialized from scratch with random weights. During training, we use the AdamW \cite{adamw} optimizer. We evaluate RAFT-Stereo after 32 disparity-field updates in our ablation experiments and after 80 updates in table \ref{table:GenResults}. \smallskip\\
\noindent\textbf{Training Schedule:} Final models are trained on synthetic data for 200k steps with a batch size of 8, while ablation experiments are trained with a batch size of 6 for 100k steps. Ablation experiments (see table \ref{table:Ablations}) are run with 16 disparity-field updates during training, and final results were trained with 22 updates. We use a one-cycle learning rate schedule \cite{smith2018superconvergence} with a minimum learning rate of $1e^{-4}$. All RAFT-Stereo experiments were trained on random $360$x$720$ crops (excluding benchmark submissions) and all experiments, excluding ablation experiments, were trained using data augmentation. Specifically: the image saturation was adjusted between 0 (greyscale) and 1.4; the right image was perturbed to simulate imperfect rectification that is common in datasets such as ETH3D and Middlebury; we stretch the images and disparity by random factors in the range $[2^{-0.2}, 2^{0.4}]$ in order to simulate a range of possible disparity distributions. 

\subsection{Zero-Shot Generalization}
\setlength\tabcolsep{.4em}
\begin{table}[t]
\centering
\resizebox{0.9\columnwidth}{!}{
\begin{tabular}{ lccccc }
\toprule
\multirow{2}{*}{Method} & \multirow{2}{*}{KITTI-15} & \multicolumn{3}{c}{Middlebury} & \multirow{2}{*}{ETH3D} \\
& & full & half & quarter & \\
\midrule
HD$^3$ \cite{hd3}  & 26.5 & 50.3 & 37.9 & 20.3 & 54.2 \\ 
gwcnet \cite{gwcnet} & 22.7 & 47.1 & 34.2 & 18.1 & 30.1\\ 
PSMNet \cite{psmnet} & 16.3 & 39.5 & 25.1 & 14.2 & 23.8\\ 
GANet \cite{ganet} & 11.7 & 32.2 & 20.3 & 11.2 & 14.1 \\ 
DSMNet \cite{dsmnet} & \underline{6.5} & \underline{21.8} & \underline{13.8} & \textbf{8.1} & \underline{6.2}\\ 
\midrule
Ours & \textbf{5.74} & \textbf{18.33} & \textbf{12.59} & \underline{9.36} & \textbf{3.28}\\ 
\midrule
\end{tabular}
}
\vspace{1mm}
\caption{Synthetic to real generalization experiments. All methods were trained on SceneFlow\cite{sceneflow} and tested on the KITTI-2015, Middlebury, and ETH3D validation datasets. We report average results across six independent training runs evaluated after 200k steps. Errors are the percent of pixels with end-point-error greater than the specified threshold. We use the standard evaluation thresholds: 3px for KITTI, 2px for Middlebury, 1px for ETH3D.}\vspace{-4mm}
\label{table:GenResults}
\end{table}

We evaluate RAFT-Stereo's ability to generalize from synthetic training data to unseen real-world datasets. This ability is critical as there exist no large-scale real-world datasets for training. In table \ref{table:GenResults}, we report RAFT-Stereo's generalization from Sceneflow \cite{sceneflow} directly to the KITTI-15, ETH3D and Middlebury validation sets, and compare to other methods in the same zero-shot setting.

Across all three validation datasets, RAFT-Stereo exhibits state-of-the-art performance in the zero-shot synthetic-to-real setting. RAFT-Stereo is trained for 200k iterations using data augmentation. 

\subsection{KITTI}
\setlength\tabcolsep{.4em}
\begin{table}[t]
\centering
\resizebox{0.8\columnwidth}{!}{
\begin{tabular}{lccc}
\toprule
Method & $\ \ \ $ all $ \ \ \ $ & foregr. & backgr. \\
\midrule
AcfNet \cite{zhang2020adaptive} & 1.89 & 3.80 & 1.51 \\
AMNet \cite{du2019amnet} & 1.84 & 3.43 & 1.53 \\
OptStereo \cite{wang2021pvstereo} & 1.82 & 3.43 & 1.50 \\
GANet-deep \cite{ganet} & 1.81 & 3.46 & 1.48 \\
SUW-Stereo \cite{ren2020suw} & 1.80 & 3.45 & \underline{1.47}\\
GANet + DSMNet \cite{dsmnet} & 1.77 & 3.23 & 1.48 \\
CSPN \cite{cheng2019learning} & \underline{1.74} & \textbf{2.88} & 1.51 \\
LEAStereo \cite{cheng2020hierarchical} & \textbf{1.65} & 2.91 & \textbf{1.40} \\

\midrule
Ours & 1.96 & \underline{2.89} & 1.75 \\
\midrule
\end{tabular}

}
\caption{Results on the KITTI-2015 \cite{kitti} leaderboard. Only published results are included. Best results for each evaluation metric are bolded, second best are underlined. At the time of submission, RAFT-Stereo ranks second on the percentage of erroneous (EPE $> 3.0$ px) foreground pixels among published methods. }
\label{table:KITTIResults}
\end{table}

We submit RAFT-Stereo to the KITTI-2015 stereo benchmark~\cite{kitti}.
At the time of writing this paper, RAFT-Stereo ranks second on the percentage of erroneous foreground pixels on the KITTI-2015 Stereo leaderboard (See table \ref{table:KITTIResults}) among published methods. For the KITTI leaderboard, we fine-tuned our method for 5k iterations on the KITTI training set using 320x1000 random crops, a minimum learning rate of $1e^{-5}$, and data augmentation. 
\subsection{ETH3D}
\setlength\tabcolsep{.4em}
\begin{table}[bp]
\centering
\resizebox{.99\columnwidth}{!}{
\begin{tabular}{lcccc}
\toprule
Method & {bad 0.5 (\%)} & {bad 1.0 (\%)} & {bad 2.0 (\%)} & {AvgErr} \\
\midrule
HSM\cite{yang2019hierarchical} & 10.88 & 4.00 &	1.36 & 0.28 \\
NOSS-ROB \cite{noss_rob} & 10.99 & 3.30 & 1.29 & 0.31\\
iResNet\cite{pang2017cascade} & 10.26  & 3.68 & 1.00 & 0.24 \\
AdaStereo \cite{song2020adastereo} & 10.22 & 3.09 & 0.65 & 0.24 \\
HIT-Net \cite{tankovich2020hitnet} & 7.83 & 2.79 & 0.80 & 0.20 \\
\midrule
Ours	& \textbf{7.04} & \textbf{2.44} & \textbf{0.44} & \textbf{0.18} \\
\midrule
\end{tabular}
}
\vspace{1mm}
\caption{Results on the ETH3D test set leaderboard. At the time of submission, RAFT-Stereo ranks first across every evaluation metric among all published methods. For all metrics, lower is better.}
\label{table:ETH3DResults}
\end{table}

 The ETH3D dataset is too small for training, so we directly evaluate our model trained on the SceneFlow dataset. To generalize from Sceneflow to ETH3D, we simulate ETH3D's image distribution by fine tuning the network on additional greyscale Sceneflow images with gamma adjustment to simulate the often-overexposed black-and-white images in ETH3D.  On the validation set, we note the accuracy increase from applying a large number of GRU iterations which can be performed without additional memory cost. To obtain our final validation results in table \ref{table:GenResults} and in table \ref{table:ETH3DResults}, we run RAFT-Stereo for 80 iterations. We show qualitative results on ETH3D in Fig. \ref{fig:ETHQual}. Using only synthetic training data, RAFT-Stereo ranks 1st on the ETH3D two-view stereo leaderboard\cite{eth3d} among published methods, achieving a bad 1-pixel error (\% of pixels with end-point-errors greater than 1px) of 2.44, outperforming the next best result of 2.69 by 9.3\%.

\subsection{Middlebury}

\setlength\tabcolsep{.4em}
\begin{table*}[bp]
\centering
\resizebox{0.6\linewidth}{!}{
\begin{tabular}{lcccccc}
\toprule
{Methods} & {AvgErr} & {MedErr} & {bad 0.5 (\%)} & {bad 1.0 (\%)} & {bad 2.0 (\%)} & {bad 4.0 (\%)} \\
\midrule
EdgeStereo \cite{edgestereo} & 2.68 & 0.72 & 55.6 & 32.4 & 18.7 & 10.8\\
HSM-Net \cite{hsmnet} & 2.07 & 0.56 & 50.7 & 24.6 & 10.2 & 4.83\\
LEAStereo \cite{cheng2020hierarchical} & 1.43 & 0.53 & 49.5 & 20.8 & 7.15 & \textbf{2.75}\\
MC-CNN \cite{mccnn} & 2.63 & 0.44 & 40.1 & 16.1 & 6.35 & 3.81\\
LocalExp \cite{localexp} & 2.24 & 0.43 & 38.7 & 13.9 & 5.43 & 3.69\\
CRLE \cite{crle} & 2.25 & 0.42 & 38.1& 13.4 & 5.75 & 3.90\\
HITNet \cite{tankovich2020hitnet} & 1.71 & 0.40 & 34.2 & 13.3 & 6.46 & 3.81\\
NOSS-ROB \cite{noss_rob} & 2.08 & 0.42 & 38.2 & 13.2 & 5.01 & 3.46\\
\midrule
Ours & \textbf{1.27} & \textbf{0.26} & \textbf{27.7} & \textbf{9.37} & \textbf{4.74} & \textbf{2.75}\\
\midrule
\end{tabular}
}
\vspace{0.1cm}\caption{Results on the Middlebury test set leaderboard compared to the top performing methods. Lower is better for all metrics.}
\label{table:MiddleburyResults}
\end{table*}

\begin{figure*}[t]
\centering
	\includegraphics[width=.99\textwidth]{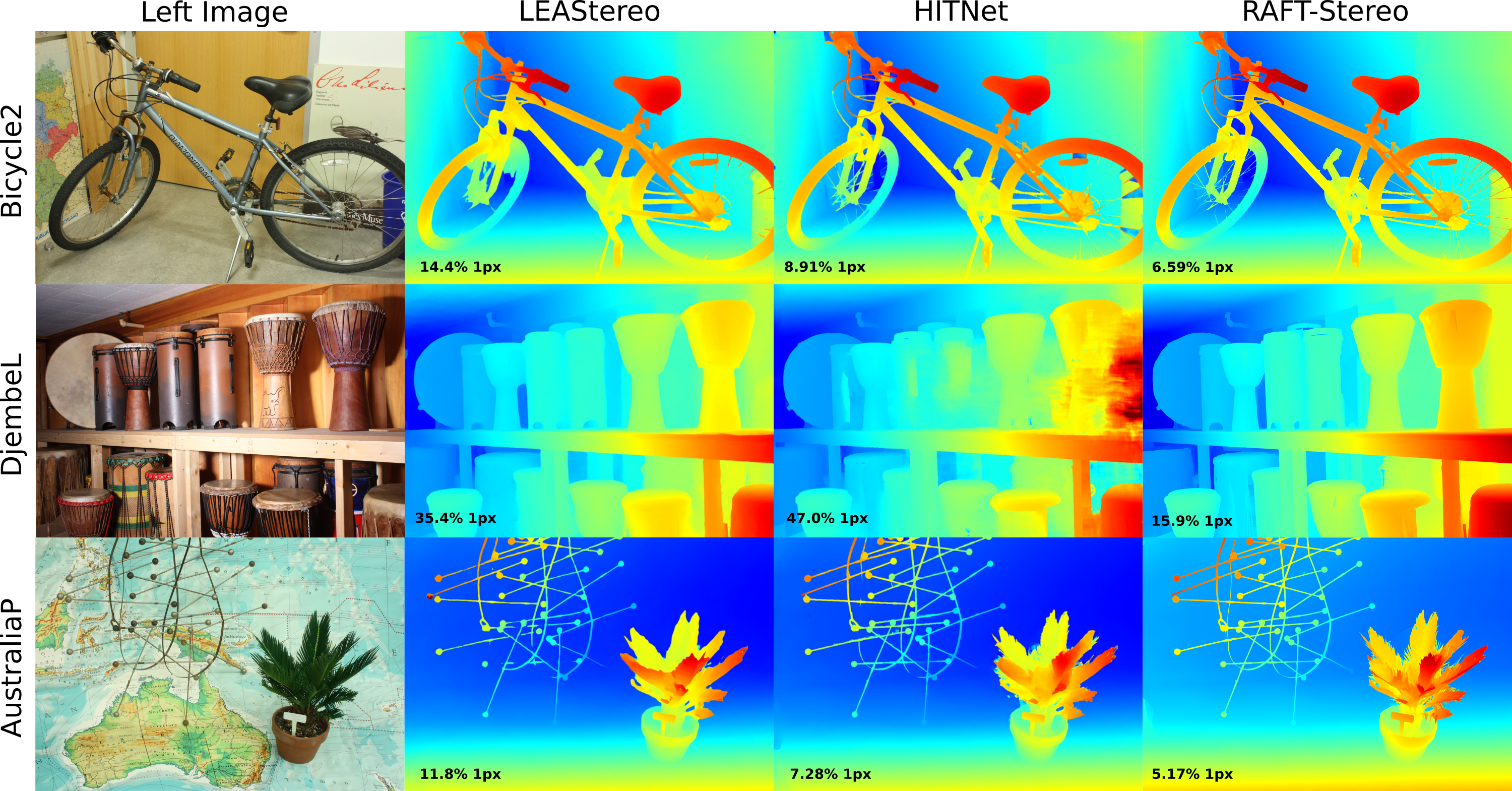}
	\caption{Results on the Middlebury \cite{middlebury} test set compared to the top end-to-end deep learning approaches. We also report the 1px error of each output in the corner. RAFT-Stereo is able to recover extremely fine details that other approaches cannot, such as the spokes of the bike wheel, the individual leaves of the plant, and sharp object boundaries.}
	\label{fig:MBQual}
\end{figure*}

RAFT-Stereo ranks first on the Middlebury Test set leaderboard, with a bad 2px error of 4.74\%, a 26\% reduction in error over the next best end-to-end deep learning method. See table \ref{table:MiddleburyResults}. The Middlebury dataset provides 23 high resolution image pairs for training and/or validation, as well as versions with alternate lighting. After pre-training on Sceneflow \cite{sceneflow}, we fine-tune on 384x1000 random crops of the 23 Middlebury traning images for 4000 steps with a batch size of 2, using 22 update iterations during training, and 32 at inference.

RAFT-Stereo is extremely memory efficient, and is therefore able to output full-resolution ($1900$x$3000$) dense optical flow. This is in contrast to 33 of the remaining 34 best methods on the leaderboard, which require upsampling their output from half-resolution. To further reduce memory, we also adapt RAFT's memory efficient correlation implementation to 3D, where correlation features are computed on-the-fly. We refer the reader to section 3.2 in RAFT \cite{teed2020raft} for more information. Beyond the aforementioned horizontal image stretching, saturation adjustment and vertical perturbation of the right image, we do no additional data augmentation to adapt to the Middlebury dataset. Fig. \ref{fig:MBQual} shows qualitative results of RAFT-Stereo on Middlebury.

\subsection{Synthetic Datasets}

\begin{table*}[t]
\centering
\resizebox{0.8\textwidth}{!}{
\begin{tabular}{cccc|ccc}
\toprule
{Sceneflow \cite{sceneflow}} & {Falling Things \cite{DBLP:journals/corr/abs-1804-06534}} & {Tartan Air \cite{tartanair2020iros}} & {Sintel Stereo \cite{Butler:ECCV:2012}}  & {ETH3D} & {KITTI-15} & {Middlebury (Full)} \\
\midrule
\checkmark  & - & - & - & \textbf{4.44} & {6.37} & {23.40}\\
-  & \checkmark & - & - & {26.93} & {6.13} & {25.39}\\ 
- & - & \checkmark & - & {4.62} & 5.87 & {26.28}\\ 
\checkmark  & \checkmark & - & - & {25.2} & {5.88} & 20.95\\ 
\checkmark  & \checkmark & \checkmark & - & {7.65} & {5.76} & \textbf{20.65}\\ 
\checkmark &  \checkmark &  \checkmark &  \checkmark & {5.65} & \textbf{5.62} & {21.99}\\ 
\midrule
\end{tabular}
}
\caption{Synthetic data generalization experiments. All experiments were run twice with different weight initializations and the validation performances were averaged. Data were balanced so that each dataset represents an equal proportion of the training data. Experiments were done using RAFT-Stereo with a single hidden-state with random cropping and vertical perturbation of the right image. }
\label{table:syntheticdata}
\end{table*}

In order to improve zero-shot generalization performance, we train additional versions of RAFT-Stereo using additional synthetic data. As real-world stereo correspondence training data is difficult to obtain en masse, most stereo correspondence works such as PSMNet \cite{psmnet} and DSMNet \cite{dsmnet} leverage synthetic training data, specifically only the Sceneflow dataset, for training.

The overall structure of the 3D scenes in Sceneflow, however, is not representative of other real-world datasets to which we hope to generalize. To remedy this, we investigate three additional publicly available synthetic datasets and demonstrate that combining them with Sceneflow can improve zero-shot generalization performance. We show in table \ref{table:syntheticdata} that certain combinations of datasets benefit generalization to specific validation datasets. \smallskip\\ 
\noindent\textbf{Falling Things:} Falling things \cite{DBLP:journals/corr/abs-1804-06534} is a photo-realistic synthetic dataset of miscellaneous objects placed sporadically around a scene. Originally intended as an object detection and 3D pose estimation dataset, Falling Things provides 61.5K image pairs for training stereo correspondence methods. We demonstrate that the use of this dataset improves generalization performance, specifically to the KITTI and Middlebury datasets.\smallskip\\
\noindent\textbf{Tartan Air:} Tartan Air \cite{tartanair2020iros} is a publicly available photo-realistic synthetic dataset of simulation environments modeled after real-world settings. This dataset was intended primarily as a SLAM dataset, but also provides 296K image pairs for training stereo correspondence methods. In our experiments, we show that Tartan Air generalizes well to KITTI and to ETH3D. \smallskip\\
\noindent\textbf{Sintel-Stereo:} The Sintel dataset \cite{Butler:ECCV:2012} exists primarily as a synthetic dataset for training optical flow methods. In addition to optical flow training data, they also provide 2.1K image pairs for training stereo correspondence methods. While training a RAFT-Stereo exclusively on this dataset caused it to overfit, we found that leveraging Sintel-Stereo together with all three other synthetic datasets gave excellent generalization performance, specifically in that it improves generalization to the ETH3D dataset.

\subsection{Ablations}
\label{sec:ablations}

\begin{figure}[ht]
    \centering
    \vspace{-8mm}
	\includegraphics[width=.99\columnwidth]{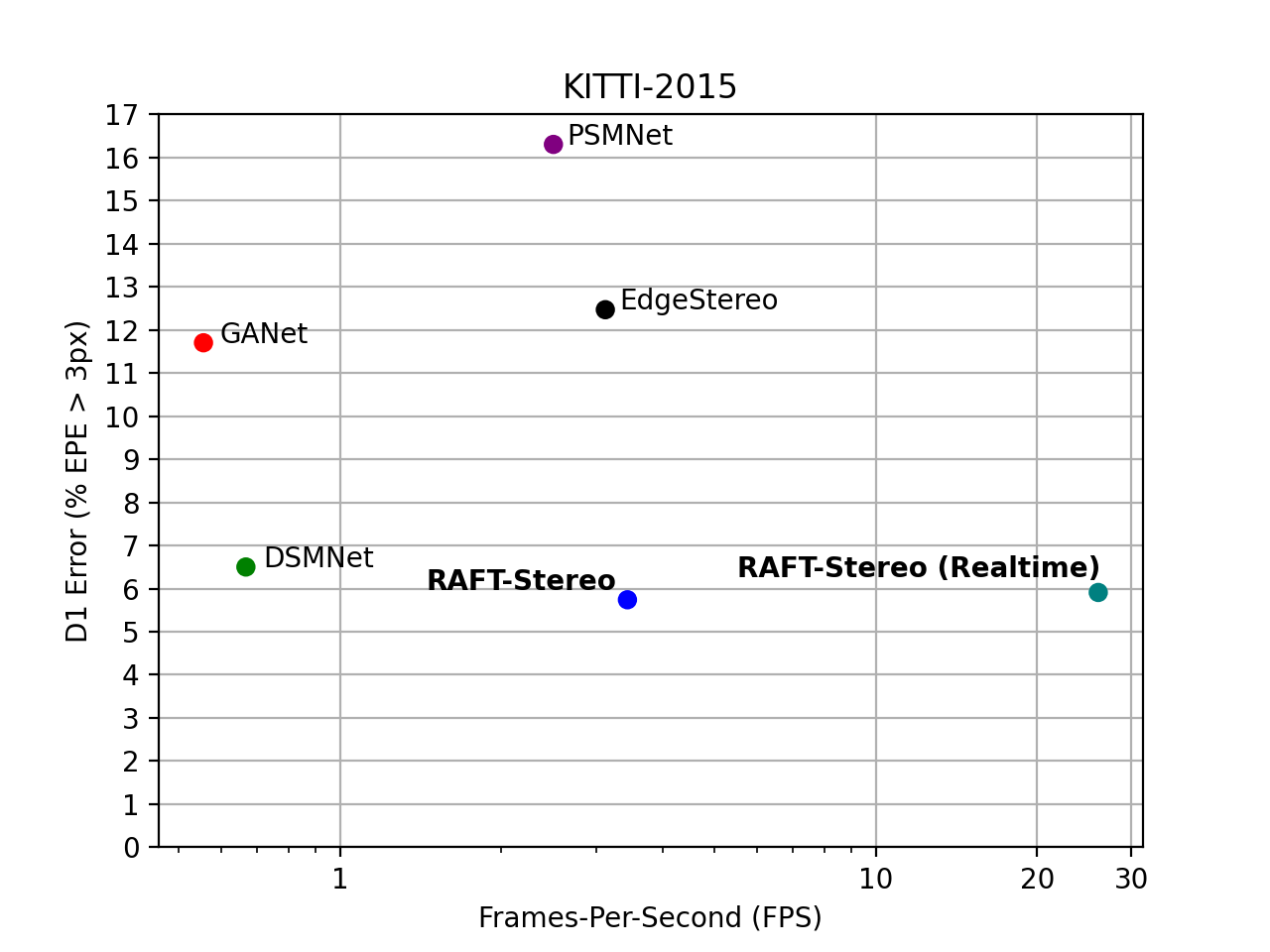}
	\caption{Plot comparing zero-shot generalization from synthetic data to KITTI-2015. All methods are trained only on Sceneflow \cite{sceneflow}, without any fine-tuning. RAFT-Stereo can be configured for real time inference and achieves competitive performance with the state-of-the-art stereo methods. Relative to our base model (blue), our realtime model (teal)  uses a shared backbone, two hidden state resolutions and slow-fast GRUs updating the flow-field at 1/8th resolution (Sec. ~\ref{sec:realtimeinf})}
	\label{fig:Runtimes}
\end{figure}

\setlength\tabcolsep{.7em}
\begin{table*}[t]
\centering
\resizebox{0.75\textwidth}{!}{
\begin{tabular}{clccc}
\toprule
\multirow{2}{*}{Experiment} & \multirow{2}{*}{Method} & \multirow{2}{*}{FlyingThings3D} & \multirow{2}{*}{Runtime (s)} & \multirow{2}{*}{Parameters} \\
\\
\midrule 
\multirow{2}{*}{\# GRU Levels.}  & \underline{3 Levels} & \textbf{9.40} & 0.132 & 11.23M \\
                            & 1 Level & 9.64 & \textbf{0.091} & 9.46M \\ \midrule
\multirow{2}{*}{Backbone}      & Single Backbone & \textbf{9.37} & \textbf{0.121} & 10.75M \\
                            & \underline{Sep. Backbones} & 9.40 & 0.132 & 11.23M\\ \midrule
\multirow{2}{*}{Resolution}    & \underline{1/4th} & \textbf{7.92} & 0.338 & 11.12M \\
                            & 1/8th & 9.40 & \textbf{0.132} & 11.23M\\ \midrule
\multirow{2}{*}{Slow-Fast GRU} & \underline{Regular} & \textbf{9.40} & 0.132 & 11.23M \\
                            & Slow-Fast & {9.98} & \textbf{0.063} & 11.23M\\ \midrule
\multirow{2}{*}{Collapsed Cost Volume} & RAFT & - & 0.224 & 5.26M \\
                            & RAFT-Stereo & - & \textbf{0.132} & 11.23M\\ \midrule
\end{tabular}
}
\caption{Ablation experiments. Settings used in our final model are underlined. See section \ref{sec:ablations} for details. All experiments are run for 100k steps on random 320x720 crops of Sceneflow with vertical perturbations to the right image as the only augmentation. Methods are evaluated on the held-out FlyingThings3D \cite{sceneflow} test set which we used to make all design decisions for our zero-shot generalization / real-time experiments. We report the 1px error evaluated after 100k steps, averaged across two independent training runs with random weight initialization. The reported the runtime comparisons were made on 1248x384 resolution images (i.e. KITTI resolution).}
\label{table:Ablations}
\end{table*}
\noindent
\textbf{GRU Levels:} RAFT-Stereo maintains and updates multiple hidden states at multiple resolutions, typically 1/8, 1/16 and 1/32 resolutions as show in figure  \ref{fig:multileveldiagram}. Each hidden state is updated using a dedicated GRU which uses the adjacent hidden states as context in addition to specific context features for that resolution. Using multiple hidden states increases the runtime but results in better performance overall. \smallskip\\
\noindent\textbf{Backbone:} RAFT-Stereo uses separate backbones in order to extract correlation features and context features for the GRU updates. We show that using a single backbone to produce both the correlation features and context features leads to faster inference without incurring any decrease in performance. We use a single-backbone architecture in the real-time version of RAFT (See section \ref{sec:realtimeinf} and Fig. \ref{fig:Runtimes}).\smallskip\\
\noindent\textbf{Resolution:} RAFT-Stereo updates its running estimate of the disparity at 1/8 or 1/4 resolution. Maintaining the running disparity estimate at 1/4 resolution yields significantly better generalization, but results in slower runtimes and uses approximately 4x as much GPU memory. This is done by shrinking the stride in the feature extractors and by predicting a proportionally smaller mask for convex upsampling.\smallskip\\
\noindent\textbf{Collapsed Cost Volume:} Rather than training a separate network for estimating stereo correspondence, one option is to apply an existing optical flow method and project the predicted flow onto the epipolar line. We show that specializing RAFT for stereo by simply collapsing the cost volume gives significantly faster runtime relative to RAFT.  \smallskip\\
\noindent\textbf{Slow-Fast:} We observe a significant decrease in runtime of RAFT-Stereo by iterating the lower resolution GRUs more often and the higher resolution GRUs less often, with a limited penalty to accuracy. In table \ref{table:Ablations}, the "Slow-Fast" version of RAFT-Stereo updates the lowest, middle and highest resolution hidden states 30, 20 and 10 times, respectively, while "Regular" updates each hidden state 32 times. Both "Slow-Fast" and "Regular" use the same model weights

\subsection{Real-time Inference}
\label{sec:realtimeinf}
We demonstrate that RAFT-Stereo can be configured to achieve real-time inference on KITTI-resolution (1248x384) images with competitive performance. By leveraging Slow-Fast bi-level (1/8 and 1/16 resolution) GRUs and a single backbone, RAFT-Stereo runs at 26 FPS. Our real-time implementation of RAFT-Stereo's performance (5.91 D1 error) is competitive with DSMNet \cite{dsmnet} (6.5 D1 error). See Fig. \ref{fig:Runtimes}. Additionally, we implement our own bilinear sampler in CUDA as Pytorch's default implementation proved to be a runtime bottleneck. 
\section{Conclusions}
We have proposed RAFT-Stereo, a new deep architecture for two-view Stereo based on RAFT ~\cite{teed2020raft}. RAFT-Stereo extends RAFT by leveraging multi-level GRUs to efficiently pass information across the image. Our approach achieves state-of-the-art cross-dataset generalization and ranks first on the Middlebury benchmark and outperforms all published work on ETH3D.

\noindent \textbf{Acknowledgements} This work is partially supported by the National Science Foundation under Award IIS-1942981.


{\small
\bibliographystyle{ieee_fullname}
\bibliography{egbib}
}

\end{document}